\documentclass[letterpaper, 10 pt, conference]{ieeeconf}  % Comment this line out if you need a4paper

\IEEEoverridecommandlockouts                              % This command is only needed if 
\usepackage{cite}
\usepackage{amsmath,amssymb,amsfonts}
\usepackage[linesnumbered,ruled,vlined]{algorithm2e}
\SetKwInput{KwIn}{Input}
\SetKwInput{KwOut}{Output}
\usepackage{graphicx}
\graphicspath{{FIG/}}   % 图放 FIG/ 文件夹
\usepackage{textcomp}
\usepackage{xcolor}
\usepackage{booktabs}
\usepackage{multirow}
\usepackage{array}
\title{\LARGE \bf
Predictive Relative-Velocity Steering for Safe Robotic Manipulator Teleoperation in Dynamic Environments
}

\author{Changhao Hu$^{1,2*}$, Zeyi Liu$^{2,3*}$, Songqiao Hu$^{2,3*}$, Shuang Liu$^{2,4}$, Zihan Meng$^{4}$, and Xiao He$^{2,3\dagger}$% <-this % stops a space
\thanks{*These authors contributed equally to this work.}%
\thanks{$^{\dagger}$Corresponding author (email: hexiao@tsinghua.edu.cn).}
\thanks{$^{1}$Zhejiang University-University of Illinois Urbana-Champaign Institute, Zhejiang University, Haining 314400, China.}%
\thanks{$^{2}$Department of Automation, Tsinghua University, Beijing 100084, China.}%
\thanks{$^{3}$Institute for Embodied Intelligence and Robotics, Tsinghua University, Beijing 100084, China.}%
\thanks{$^{4}$TetraBOT.}%
\thanks{This work was supported by National Natural Science Foundation of China under grants 62525308, 62473223, and Beijing Natural Science Foundation under grant L241016.}%
}

\begin{document}

\maketitle
\thispagestyle{empty}
\pagestyle{empty}

%%%%%%%%%%%%%%%%%%%%%%%%%%%%%%%%%%%%%%%%%%%%%%%%%%%%%%%%%%%%%%%%%%%%%%%%%%%%%%%%
\begin{abstract}
Recent advances in teleoperation have enabled robotic manipulators to perform dexterous, human-arm-like motions. However, human operators may fail to avoid suddenly appearing obstacles promptly and effectively, particularly under network latency or limited attention, thereby creating safety risks. To address this issue, we propose a lightweight and modular framework for proactive collision avoidance, operating directly at the end-effector velocity-command level. After preprocessing the point cloud, the framework first predicts potential collisions based on time-to-collision (TTC) with integrated overshoot protection, and subsequently rotates the relative-velocity vector using Rodrigues' rotation formula. The deflection changes only the direction of the relative velocity while preserving its magnitude, thereby mitigating the deadlock problem commonly encountered by conventional artificial potential field (APF) methods. The prediction module compensates for point-cloud processing latency introduced by complex teleoperation pipelines, while the lightweight design enables the high-frequency control required for teleoperation. Simulations across diverse scenarios show that the proposed method achieves a higher end-effector collision avoidance rate than the baseline methods. Experiments on a physical robotic system further validate its collision-avoidance effectiveness.
\end{abstract}

\section{Introduction}

Teleoperation has emerged as a widely used paradigm for robotic manipulation because of its precision and remote operability~\cite{hokayem2006bilateral}. This paradigm enables complex and dexterous operations that remain challenging for conventional motion-planning and vision-language-action (VLA) methods, while distancing operators from hazardous environments.

Safety is a critical requirement in teleoperation~\cite{zhong2024attentiveness, he2024real}. In principle, human operators can manually issue commands to avoid obstacles. However, operators concentrating on the manipulation task may not notice obstacles in the remote environment promptly, especially when the workspace is non-stationary and obstacles may appear or move unexpectedly. Even when an operator initiates an evasive maneuver, network latency and the complex processing pipeline inherent in teleoperation may prevent the corresponding command from being executed in time. Existing studies have therefore introduced safety-assistance mechanisms for teleoperation from several directions. These mechanisms include reactive repulsion-based collision avoidance methods typified by APFs~\cite{gottardi2022shared, khatib1986real, qin2024shared}, virtual fixtures (VFs) and haptic guidance that constrain the motion trajectory~\cite{rosenberg1993virtual, pruks2022method}, and control barrier function (CBF)-based safety filters that project operator commands onto a safe set~\cite{10611280, hu2026safelinksafetycriticalcontroldynamic}. In addition, standards-based approaches such as speed and separation monitoring (SSM) in human--robot collaboration provide guidance for designing safety-distance policies~\cite{marvel2017implementing}.

However, in classical APF, attractive and repulsive forces may nearly cancel each other, resulting in local minima or velocity stagnation when the resultant force becomes small\cite{131810}. In teleoperation, an operator may misinterpret such stagnation as a network disconnection or hardware failure and consequently apply abnormal joystick commands, further increasing the risk of collision. Although VFs can provide effective guidance, they often rely on predefined fixtures or interactive construction and therefore exhibit limited robustness in unfamiliar environments. CBFs can offer stronger safety guarantees but typically require smooth obstacle representations and online optimization. Applying them directly to point-cloud inputs may thus require restrictive assumptions and incur a substantial computational burden.
\begin{figure}[t]
    \centering
    \includegraphics[width=\columnwidth]{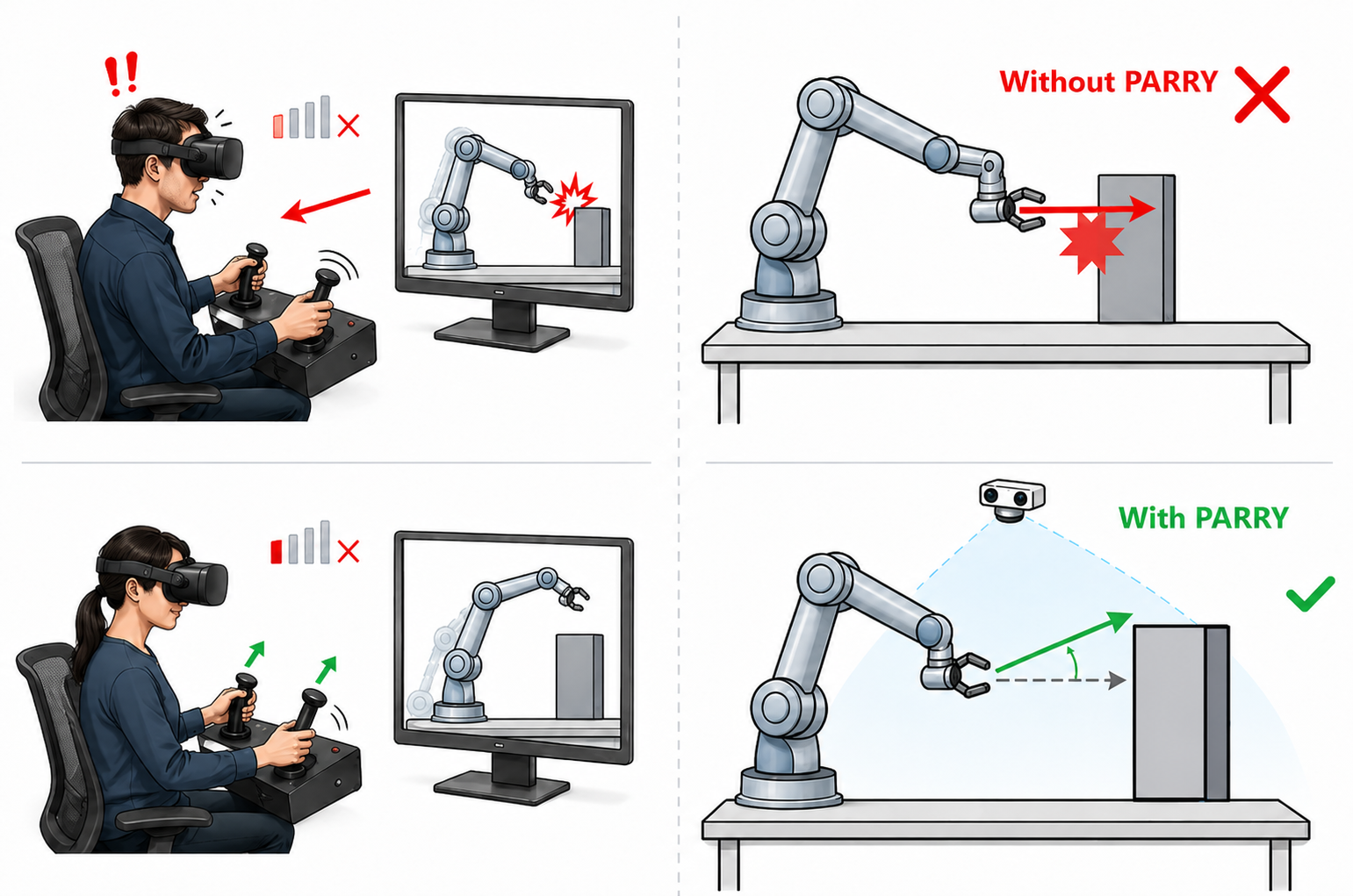}
    \caption{
    Illustration of PARRY in teleoperation.
    \textbf{Prompt:} A 2x2 scientific illustration. Top row: Male VR operator faces network lag, causing a robotic arm to crash into a grey pillar (red explosion). Bottom row: Female VR operator works smoothly, robotic arm uses overhead camera vision to deflect its path along a green curve, successfully avoiding the pillar.
    }
    \label{fig:teaser_intro}
    \vspace{-6mm}
\end{figure}

To mitigate these limitations, this paper presents PARRY (\textbf{P}redictive \textbf{A}voidance by \textbf{R}otating \textbf{R}elative velocit\textbf{Y}). The overall effect is illustrated in Fig.~\ref{fig:teaser_intro}. PARRY operates at the teleoperation velocity-command level: given a depth point cloud and a nominal end-effector velocity, it outputs a new execution velocity without modifying the underlying tracking controller. Specifically, PARRY performs TTC-aware point-cloud prediction based on the relative motion of obstacles and computes repulsive guidance from the predicted point cloud. It then applies Rodrigues' rotation formula to redirect the relative velocity without changing its magnitude. This directional deflection steers the end effector around obstacles while avoiding the velocity stagnation caused by near-zero resultant forces in classical APF methods. By directly processing point clouds through a lightweight pipeline, PARRY meets the real-time requirements of teleoperation.

The main contributions of this work are as follows.
\begin{enumerate}
\item We propose PARRY, a plug-and-play reactive collision avoidance framework for teleoperation. Operating at the velocity-command level directly, it applies Rodrigues’ rotation to redirect relative velocity, steering the end-effector around obstacles while preserving the magnitude of the relative velocity.

\item We develop a TTC-based prediction mechanism to compensate for obstacle-observation latency. The mechanism reconstructs a look-ahead point cloud and incorporates overshoot protection to prevent predicted points from passing beyond the end effector. This design enables reliable collision avoidance despite latency in the perception pipeline.

\item We conduct 1,000 paired Monte Carlo simulations under 0, 100, and 150 ms latencies, comparing PARRY with representative classical and advanced safety baselines alongside ablation studies, resulting in a total of 21,000 runs, and validate its effectiveness and feasibility on a physical 7-DoF setup.
\end{enumerate}
\section{Related Work}

\subsection{Rotational Obstacle Avoidance}

To mitigate APF-induced stagnation and oscillation, existing methods guide robots around obstacles using tangential forces or directional modulation. Singh et al. introduced the circulatory-field method~\cite{singh1996real}. Becker et al. subsequently proposed CFP and ICF, extending this approach to predictive multi-agent settings, point-cloud inputs, and whole-body collision avoidance~\cite{becker2021circular, becker2023informed}. Ataka et al. developed a magnetic-field-inspired navigation method that relies only on local sensing~\cite{ataka2022magnetic}, while Yang et al. combined magnetic-field potentials with dynamic movement primitives (DMPs) to generate trajectories around volumetric obstacles~\cite{yang2025optimization}. Souza et al. instead superimpose a vortex field obtained by rotating each obstacle's repulsive field~\cite{souza2022modified}. Khansari-Zadeh and Billard proposed a real-time dynamical-system approach that modifies the original dynamics through a state-dependent modulation matrix while preserving its attractor for convex obstacles~\cite{khansari2012dynamical}. ROAM rotates the initial dynamics toward the obstacle tangent space to handle concave obstacles~\cite{huber2023avoidance}, whereas CARE rotates two-dimensional navigation trajectories using visually estimated repulsive forces~\cite{kim2025care}. APF-WF escapes local minima by temporarily switching to a dedicated wall-following mode~\cite{kim2025escaping}.

\subsection{Delay Compensation and Prediction in Teleoperation}

Communication and processing-pipeline delays in teleoperation can degrade system stability and transparency. Existing compensation and prediction methods primarily focus on bilateral control or command tracking.

One classical approach is the Smith predictor. It constructs dynamic models of the remote robot and its environment on the local side to estimate delay-free feedback signals, such as contact forces and positions, thereby moving the communication delay outside the closed loop. Subsequent studies have incorporated this approach into teleoperation architectures and employed techniques such as online identification to account for nonlinear and time-varying environment dynamics~\cite{smith2006smith}. Another approach is model-mediated teleoperation, which constructs a local contact model of the remote environment from remote sensory measurements and uses it to generate immediate haptic feedback on the local side. The remote robot then executes only motion or force commands consistent with this model, reducing the instability caused by direct coupling between the local and remote sides under large delays~\cite{mitra2008model}. Shared-control approaches further model the operator's motion and task objectives. They use kinematic models to propagate delayed state measurements to the robot's current time and employ model predictive control (MPC) to enforce collision avoidance~\cite{lima2023model}. Beyond command-side delay compensation, real-time safety assessment under non-stationary operating conditions has also been systematically reviewed~\cite{liu2023real}.

\section{Method}

\subsection{Problem Formulation and System Overview}

We consider a teleoperation task in which an operator specifies the desired position $P$ and orientation $W$ of a robotic manipulator using a control interface, yielding the end-effector velocity $\mathbf{v}_{ee}$. Let $\mathbf{x}_{ee}$ denote the current end-effector position in the base frame. For environmental perception, a depth camera captures a raw point cloud, which is then preprocessed to extract an obstacle point cloud $\mathcal{O}=\{\mathbf{o}_i\in\mathbb{R}^3\}_{i=1}^{N}$, which represents the obstacle locations. The obstacle velocity $\mathbf{v}_{o}$ is then estimated by frame-to-frame differencing of the most hazardous point.

\begin{figure}[t]
\centering
\includegraphics[width=\columnwidth,trim=0 480 0 0,clip]{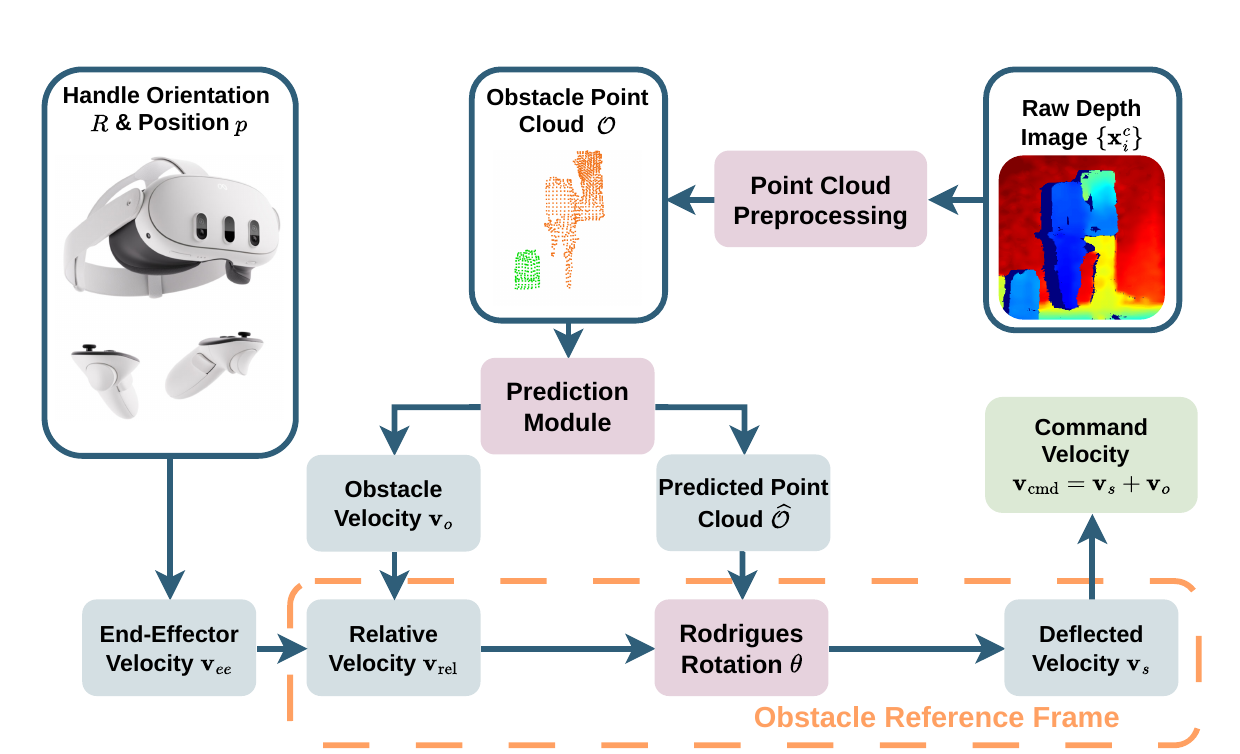}
\caption{Overall architecture of PARRY.}
\label{fig:pipeline}
\end{figure}

The end-effector velocity relative to the obstacle is defined as $\mathbf{v}_{\mathrm{rel}}=\mathbf{v}_{ee}-\mathbf{v}_{o}$. Based on the point cloud $\mathcal{O}$ and obstacle velocity $\mathbf{v}_{o}$, PARRY constructs the predicted point cloud $\widehat{\mathcal{O}}=\{\widehat{\mathbf{o}}_i\in\mathbb{R}^3\}_{i=1}^{N}$. It then computes the deflection angle $\theta$ from $\widehat{\mathcal{O}}$ and redirects $\mathbf{v}_{\mathrm{rel}}$ to obtain the modified relative velocity $\mathbf{v}_{s}$. Transforming back to the base frame gives the commanded end-effector velocity $\mathbf{v}_{cmd} = \mathbf{v}_{s} + \mathbf{v}_{o}$, which steers the manipulator around the obstacle. The overall architecture of PARRY is shown in Fig.~\ref{fig:pipeline}, and its workflow is summarized in Algorithm~\ref{alg:parry}. 

\textit{Notation:} Throughout this paper, small positive thresholds $\epsilon_a, \epsilon_o, \epsilon_f, \epsilon_v$, and $\epsilon_{\times}$ are used to ensure numerical stability and prevent computational jitter.
\begin{algorithm}[t]
\small
\caption{PARRY Velocity Deflection}
\label{alg:parry}
\KwIn{Point cloud $\mathcal{X}^{(k)}=\{\mathbf{x}_i^b\}$;
end-effector state $(\mathbf{x}_{ee}^{(k)},\mathbf{v}_{ee}^{(k)})$;
previous state $\mathcal{S}^{(k-1)}$}
\KwOut{$\mathbf{v}_{cmd}^{(k)}$ and $\mathcal{S}^{(k)}$}

\textbf{Init.:} Set $\mathbf{v}_{o}^{(0)}\leftarrow\mathbf{0}$ and choose
$\hat{\mathbf{a}}^{(0)}\perp\mathbf{v}_{ee}^{(0)}$\;

Obtain $\mathcal{O}^{(k)}$ by preprocessing $\mathcal{X}^{(k)}$\;

\eIf{$\mathcal{O}^{(k)}=\varnothing$}{
    \KwRet{$\mathbf{v}_{ee}^{(k)},\mathcal{S}^{(k-1)}$}\;
}{
    $i^\ast\leftarrow
    \underset{i}{\arg\min}\,
    \|\mathbf{o}_i^{(k)}-\mathbf{x}_{ee}^{(k)}\|_2$\;

    Estimate $\mathbf{v}_{o}^{(k)}$ from
    $\mathbf{o}_{i^\ast}^{(k)}$ and $\mathcal{S}^{(k-1)}$\;

    Generate $\widehat{\mathcal{O}}^{(k)}$ from
    $\mathcal{O}^{(k)}$, $\mathbf{x}_{ee}^{(k)}$, and
    $\mathbf{v}_{o}^{(k)}$\;

    Compute $\mathbf{F}_{\mathrm{rep}}^{(k)}$ from
    $\widehat{\mathcal{O}}^{(k)}$ and $\mathbf{x}_{ee}^{(k)}$\;

    $\mathbf{v}_{\mathrm{rel}}^{(k)}
    \leftarrow\mathbf{v}_{ee}^{(k)}-\mathbf{v}_{o}^{(k)}$\;

        \eIf{degenerate or receding}{
      keep $\mathbf{v}_{s}^{(k)}\leftarrow\mathbf{v}_{\mathrm{rel}}^{(k)}$, $\hat{\mathbf{a}}^{(k)}\leftarrow\hat{\mathbf{a}}^{(k-1)}$\;
    }{
        Determine $\hat{\mathbf{a}}^{(k)}$ from
        $\mathbf{v}_{\mathrm{rel}}^{(k)}$,
        $\mathbf{F}_{\mathrm{rep}}^{(k)}$, 
        $\hat{\mathbf{a}}^{(k-1)}$\;

        $\theta^{(k)}\leftarrow
        \operatorname{clip}\!\left(
        k_{\mathrm{rot}}\|\mathbf{F}_{\mathrm{rep}}^{(k)}\|_2,
        0,\theta_{\max}\right)$\;

        Rotate $\mathbf{v}_{\mathrm{rel}}^{(k)}$ by
        $\hat{\mathbf{a}}^{(k)}$, $\theta^{(k)}$
        to obtain $\mathbf{v}_{s}^{(k)}$\;
    }

    $\mathbf{v}_{cmd}^{(k)}
    \leftarrow\mathbf{v}_{s}^{(k)}+\mathbf{v}_{o}^{(k)}$\;

    $\mathcal{S}^{(k)}\leftarrow
    \bigl(\mathbf{o}_{i^\ast}^{(k)},
    \mathbf{v}_{o}^{(k)},\hat{\mathbf{a}}^{(k)}\bigr)$\;

    \KwRet{$\mathbf{v}_{cmd}^{(k)},\mathcal{S}^{(k)}$}\;
}
\end{algorithm}
\subsection{Self-Filtering-Based Point-Cloud Preprocessing}

Using the depth image and camera intrinsics, valid depth pixels are back-projected into 3D points $\mathbf{x}_i^c\in\mathbb{R}^3$ in the camera frame. These points are then transformed into the manipulator base frame using the hand--eye calibration:
\begin{equation}
\mathbf{x}_i^b=\mathbf{R}_{bc}\mathbf{x}_i^c+\mathbf{t}_{bc},
\end{equation}
where $\mathbf{x}_i^b$ denotes the coordinates of the same point in the base frame, and $\mathbf{R}_{bc}$ and $\mathbf{t}_{bc}$ are the rotation and translation from the camera frame to the base frame. The point cloud is subsequently cropped according to the workspace and tabletop boundaries, and voxel-downsampled using a voxel side length of $l_v$. The mean of the points within each nonempty voxel is selected as its representative point, denoted by $\mathbf{o}_i$.

Robot self-filtering is then applied to the sampled point set $\{\mathbf{o}_i\}$. The manipulator links are approximated by $M$ capsules, where the $j$th capsule is represented by endpoints $\mathbf{a}_j$ and $\mathbf{b}_j$ and radius $r_j$. Let $d_{ij}$ denote the minimum distance from $\mathbf{o}_i$ to the centerline of the $j$th capsule. Only points satisfying $d_{ij}>r_j$ for all $j=1,\dots,M$ are retained. To suppress depth noise and reduce fluctuations in the subsequent velocity estimates, radius-based outlier removal is further applied. Given a neighborhood radius $r_d$ and a minimum number of neighboring points $N_{\min}$, points with sufficiently many neighbors are retained, yielding the preprocessed point cloud
\begin{equation}
\mathcal{O}
=
\left\{
\mathbf{o}_i
\;\middle|\;
\left|
\left\{
\mathbf{o}_m:
0<\left\|\mathbf{o}_m-\mathbf{o}_i\right\|_2\le r_d
\right\}
\right|
\ge N_{\min}
\right\},
\end{equation}
where $\mathbf{o}_i$ is the sampled point being evaluated, and $\mathbf{o}_m$ denotes another sampled point within its neighborhood.

\subsection{Overshoot-Protected Adaptive Prediction}

Teleoperation pipelines typically involve multiple interconnected processes, which may cause substantial contention for hardware resources and increase the computational burden of point-cloud processing~\cite{zhang2025understanding}. For non-stationary data streams, drift detection and adaptation have been investigated in learning-based real-time safety assessment~\cite{hu2024cadm+}. In contrast, PARRY addresses the temporal mismatch caused by sensing and processing delays through geometric point-cloud prediction. To compensate for this delay, we draw on the concept of TTC~\cite{olivares2023time}. In this work, a TTC-inspired metric is defined as the obstacle's radial approach time and is used to adaptively determine the point-cloud prediction horizon.

Computationally efficient iterative updates have also been developed for online real-time safety assessment of dynamic systems~\cite{liu2024online}. To retain obstacle boundary points and keep a lightweight computation suitable for the control frequency of teleoperation, we identify the most hazardous point in the entire point cloud $\mathcal{O}$ and use its estimated obstacle velocity. For the point cloud at frame $k$, $\mathcal{O}^{(k)}=\{\mathbf{o}_i^{(k)}\}_{i=1}^{N}$, the index of the most hazardous point is defined as
\begin{equation}
i^\ast
=
\arg\min_i
\left\|
\mathbf{o}_i^{(k)}-\mathbf{x}_{ee}^{(k)}
\right\|_2.
\end{equation}

The instantaneous obstacle velocity is estimated by differencing the most hazardous points in two consecutive frames:
\begin{equation}
\tilde{\mathbf{v}}_{o}^{(k)}
=
\frac{
\mathbf{o}_{i^\ast}^{(k)}-\mathbf{o}_{i^\ast}^{(k-1)}
}{
\Delta t
},
\end{equation}
where $\Delta t$ is the sampling time step.
An exponential moving average (EMA) is applied to suppress point-cloud jitter and abrupt nearest-point switching. If the displacement of the nearest point exceeds the threshold $d_{j}$, the velocity estimate from the previous frame is retained:
\begin{equation}
\mathbf{v}_{o}^{(k)}
=
\begin{cases}
\mathbf{v}_{o}^{(k-1)},
&
\left\|
\mathbf{o}_{i^\ast}^{(k)}-\mathbf{o}_{i^\ast}^{(k-1)}
\right\|_2
>
d_{j},\\
\alpha\tilde{\mathbf{v}}_{o}^{(k)}+(1-\alpha)\mathbf{v}_{o}^{(k-1)},
&
\text{otherwise},
\end{cases}
\end{equation}
where $\alpha \in (0,1)$ is the smoothing factor.
Let $\mathbf{n}$ denote the unit vector pointing from the end effector toward the most hazardous point. Assuming a constant obstacle velocity, the radial approach speed and radial approach time are respectively defined as $v_{\mathrm{app}} = -(\mathbf{v}_{o}^{(k)})^\mathrm{T}\mathbf{n}$ and $T_{\mathrm{app}} = \|\mathbf{o}_{i^\ast}^{(k)}-\mathbf{x}_{ee}^{(k)}\|_2 / v_{\mathrm{app}}$.

The prediction horizon $t_p$ is then determined adaptively:
\begin{equation}
t_p
=
\begin{cases}
\operatorname{clip}\left(\beta\,T_{\mathrm{app}},\,t_{\min},\,t_{\max}\right),
&
v_{\mathrm{app}}>\epsilon_a,\\
t_{\min},
&
\text{otherwise},
\end{cases}
\end{equation}
where $\beta\in(0,1)$ is a scaling factor and $[t_{\min},t_{\max}]$ specifies the allowable range of the prediction horizon. 

The look-ahead displacement of the point cloud is $\Delta\mathbf{o}=\mathbf{v}_{o}^{(k)}t_p$, with its magnitude capped at $d_{p}$. Without overshoot protection, each predicted point is given by
\begin{equation}
\mathbf{o}_{i,p}^{(k)}
=
\mathbf{o}_i^{(k)}
+
\Delta\mathbf{o}.
\end{equation}

Because all points share the same velocity estimate, a predicted point may pass beyond the end effector and induce a reverse repulsive force. Alternatively, it may move farther from the end effector and thereby underestimate the collision risk. To prevent these cases, overshoot protection is applied independently to each obstacle point. Let $d_i=\|\mathbf{o}_i^{(k)}-\mathbf{x}_{ee}^{(k)}\|_2$ and $\hat{\mathbf{r}}_i=(\mathbf{o}_i^{(k)}-\mathbf{x}_{ee}^{(k)})/d_i$. Given the minimum allowable distance $d_{s}$ between a predicted point and the end effector, the predicted radial coordinate is $\ell_{i,p} = d_i+\hat{\mathbf{r}}_i^\mathrm{T}\Delta\mathbf{o}$, and the corrected point is computed as
\begin{equation}
\tilde{\mathbf{o}}_i^{(k)}
=
\mathbf{o}_i^{(k)}+\Delta\mathbf{o}
+
\left[
\min(d_i,d_{s})-\ell_{i,p}
\right]_+
\hat{\mathbf{r}}_i ,
\end{equation}
where $[z]_+=\max(z,0)$, and $\tilde{\mathbf{o}}_i^{(k)}$ denotes the predicted point after the first correction. If the corrected point remains farther from the end effector than the original point, it is reverted to the original point to avoid inadvertently introducing an erroneous reverse repulsive force:
\begin{equation}
\widehat{\mathbf{o}}_i^{(k)}
=
\begin{cases}
\tilde{\mathbf{o}}_i^{(k)},
&
\left\|\tilde{\mathbf{o}}_i^{(k)}-\mathbf{x}_{ee}^{(k)}\right\|_2<d_i+\epsilon_o,\\
\mathbf{o}_i^{(k)},
&
\text{otherwise}.
\end{cases}
\end{equation}

The final predicted point cloud is $\widehat{\mathcal{O}}=\{\widehat{\mathbf{o}}_i^{(k)}\}_{i=1}^{N}$.

\subsection{Rodrigues-Rotation-Based Safe Velocity Generation}

Inspired by the classical artificial potential field method~\cite{khatib1986real}, this section develops an end-effector collision-avoidance strategy. Given the predicted point cloud $\widehat{\mathcal{O}}=\{\widehat{\mathbf{o}}_i\}_{i=1}^{N}$, the distance between the predicted obstacle point and the end effector is $\rho_i = \|\widehat{\mathbf{o}}_i-\mathbf{x}_{ee}\|_2$. The points within the end-effector influence range form the active index set $\mathcal{I} = \{i \mid 0 < \rho_i < d_{\max}\}$.

A classical APF constructs an attractive potential toward the goal and a repulsive potential around obstacles. For the desired position $P$, the attractive potential and force are
$U_{\mathrm{att}}=\frac12\xi\|\mathbf{x}_{ee}-P\|_2^2$
and
$\mathbf{F}_{\mathrm{att}}=\xi(P-\mathbf{x}_{ee})$,
respectively, where $\xi$ is the attractive gain. Let $d_{\max}$ denote the obstacle influence distance. The classical repulsive potential associated with an individual obstacle point is
\begin{equation}
U_{\mathrm{rep},i}
=
\begin{cases}
\displaystyle
\frac{1}{2}\eta
\left(
\frac{1}{\rho_i}
-
\frac{1}{d_{\max}}
\right)^2,
&
0<\rho_i<d_{\max},\\
0,
&
\rho_i\ge d_{\max},
\end{cases}
\end{equation}
where $\eta$ is the repulsive gain. The corresponding pointwise repulsive force is
\begin{equation}
\mathbf{f}_{i}^{\mathrm{APF}}
=
\begin{cases}
\displaystyle
\eta
\left(
\dfrac{1}{\rho_i}
-
\dfrac{1}{d_{\max}}
\right)
\dfrac{1}{\rho_i^2}
\dfrac{\mathbf{x}_{ee}-\widehat{\mathbf{o}}_i}{\rho_i},
&
0<\rho_i<d_{\max},\\[0.3em]
\mathbf{0},
&
\rho_i\ge d_{\max}.
\end{cases}
\end{equation}

The aggregate repulsive force is $\mathbf{F}_{\mathrm{rep}} = \sum_{i\in\mathcal{I}}\mathbf{f}_{i}^{\mathrm{APF}}$. A classical APF then forms the resultant force $\mathbf{F}_{\mathrm{APF}} = \mathbf{F}_{\mathrm{att}} + \mathbf{F}_{\mathrm{rep}}$ and uses it to determine the collision-avoidance motion.

PARRY retains $\mathbf{F}_{\mathrm{rep}}$ as the notation for the aggregate repulsive force but computes it differently and does not directly add it to the attractive force.
Unlike methods that execute predefined trajectories, teleoperation must consider the operator's real-time commands. To provide adjustable sensitivity to obstacle distance while reducing the influence of distant obstacles on the operator's intent, PARRY adopts a power-law pointwise repulsive force, where $p>1$ is the distance-decay hyperparameter:
\begin{equation}
\mathbf{f}_i
=
-
\frac{\eta}{\rho_i^p}
\left(
\widehat{\mathbf{o}}_i-\mathbf{x}_{ee}
\right),
\qquad i\in\mathcal{I}.
\end{equation}

To prevent extensive but distant obstacle point clouds from dominating the avoidance direction, the final repulsive force $\mathbf{F}_{\mathrm{rep}}$ is computed using distance-weighted aggregation:
\begin{equation}
\mathbf{F}_{\mathrm{rep}}
=
\frac{
\sum_{i\in\mathcal{I}}w_i\mathbf{f}_i
}{
\sum_{i\in\mathcal{I}}w_i
},
\end{equation}
where $w_i = 1/\rho_i^p$ is the distance weight.
In the obstacle reference frame, the $\mathbf{v}_{\mathrm{rel}}$ defined in the problem formulation is used as the velocity before redirection.

The rotation axis is first computed as
\begin{equation}
\hat{\mathbf{a}}
=
\frac{
\mathbf{v}_{\mathrm{rel}}
\times
\mathbf{F}_{\mathrm{rep}}
}{
\left\|
\mathbf{v}_{\mathrm{rel}}
\times
\mathbf{F}_{\mathrm{rep}}
\right\|_2
}.
\end{equation}
\begin{figure}[b]
\vspace{-6mm}
\centering
\includegraphics[width=\textwidth,trim=0 630 0 30,clip]{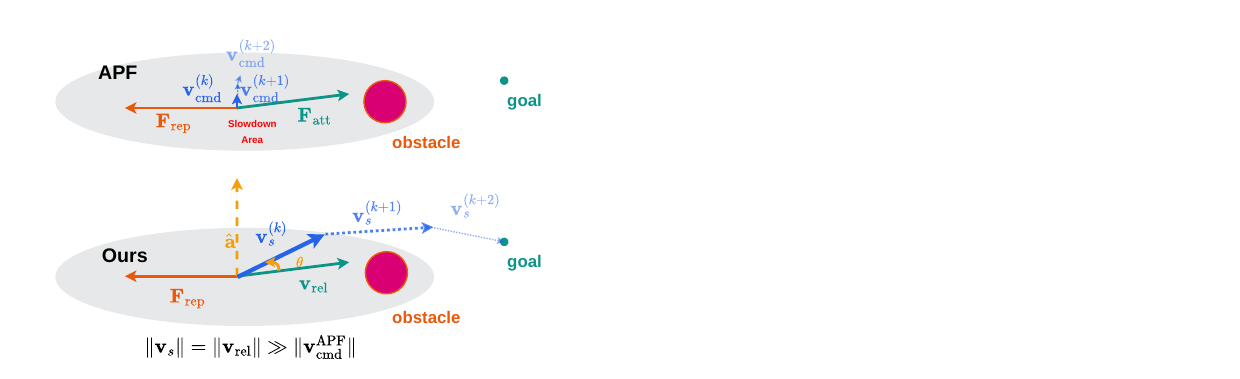}
\caption{Comparison between APF velocity superposition and PARRY relative-velocity redirection. }
\label{fig:principle}
\end{figure}

For mathematical well-posedness and control robustness, no redirection is applied---keeping the relative velocity unchanged---when $\mathcal{I}=\varnothing$, $\left\|\mathbf{F}_{\mathrm{rep}}\right\|_2\le\epsilon_f$, $\left\|\mathbf{v}_{\mathrm{rel}}\right\|_2\le\epsilon_v$, or when the end effector is already moving away from the obstacle ($\mathbf{v}_{\mathrm{rel}}^\mathrm{T}\mathbf{F}_{\mathrm{rep}}>0$ and $\left\|\mathbf{v}_{\mathrm{rel}}\times\mathbf{F}_{\mathrm{rep}}\right\|_2\le\epsilon_{\times}$). When the vectors are approximately antiparallel, the rotation axis becomes degenerate. PARRY then reuses the previous rotation axis $\hat{\mathbf{a}}^{(k-1)}$ to maintain smooth temporal continuity and stability in the redirection.

Importantly, in this antiparallel configuration, PARRY preserves the relative-velocity magnitude and therefore keeps the manipulator moving. Even if the vectors are collinear at the current time step, the continuing motion changes the robot configuration at the next time step, allowing the system to leave the degenerate state and resume normal redirection. In contrast, the resultant APF velocity approaches zero in this configuration, preventing the manipulator from adjusting its position. This transient velocity stagnation may cause the operator to misdiagnose the system state and issue abnormal commands, as illustrated in Fig.~\ref{fig:principle}.

The redirection angle is determined by the magnitude of the repulsive force:
\begin{equation}
\theta
=
\operatorname{clip}
\left(
k_{\mathrm{rot}}
\left\|
\mathbf{F}_{\mathrm{rep}}
\right\|_2,
0,
\theta_{\max}
\right),
\end{equation}
and $\theta_{\max}$ is the maximum allowable redirection angle.
Rodrigues' rotation formula is then applied directly to the original relative velocity vector:
\begin{equation}
\mathbf{v}_{s}
=
\mathbf{v}_{\mathrm{rel}}\cos\theta
+
\left(
\hat{\mathbf{a}}
\times
\mathbf{v}_{\mathrm{rel}}
\right)
\sin\theta
+
\hat{\mathbf{a}}
\left(
\hat{\mathbf{a}}^\mathrm{T}
\mathbf{v}_{\mathrm{rel}}
\right)
\left(
1-\cos\theta
\right).
\end{equation}

This mechanism combines the resistance of rotational avoidance to local minima with the maneuverability required for teleoperation in unstructured dynamic environments. Finally, transforming back to the base frame yields the commanded end-effector velocity:
\begin{equation}
\mathbf{v}_{\mathrm{cmd}}
=
\mathbf{v}_{s}
+
\mathbf{v}_{o}.
\end{equation}
\section{Simulation Experiments}

\begin{figure}[b]
\centering
\includegraphics[width=\columnwidth]{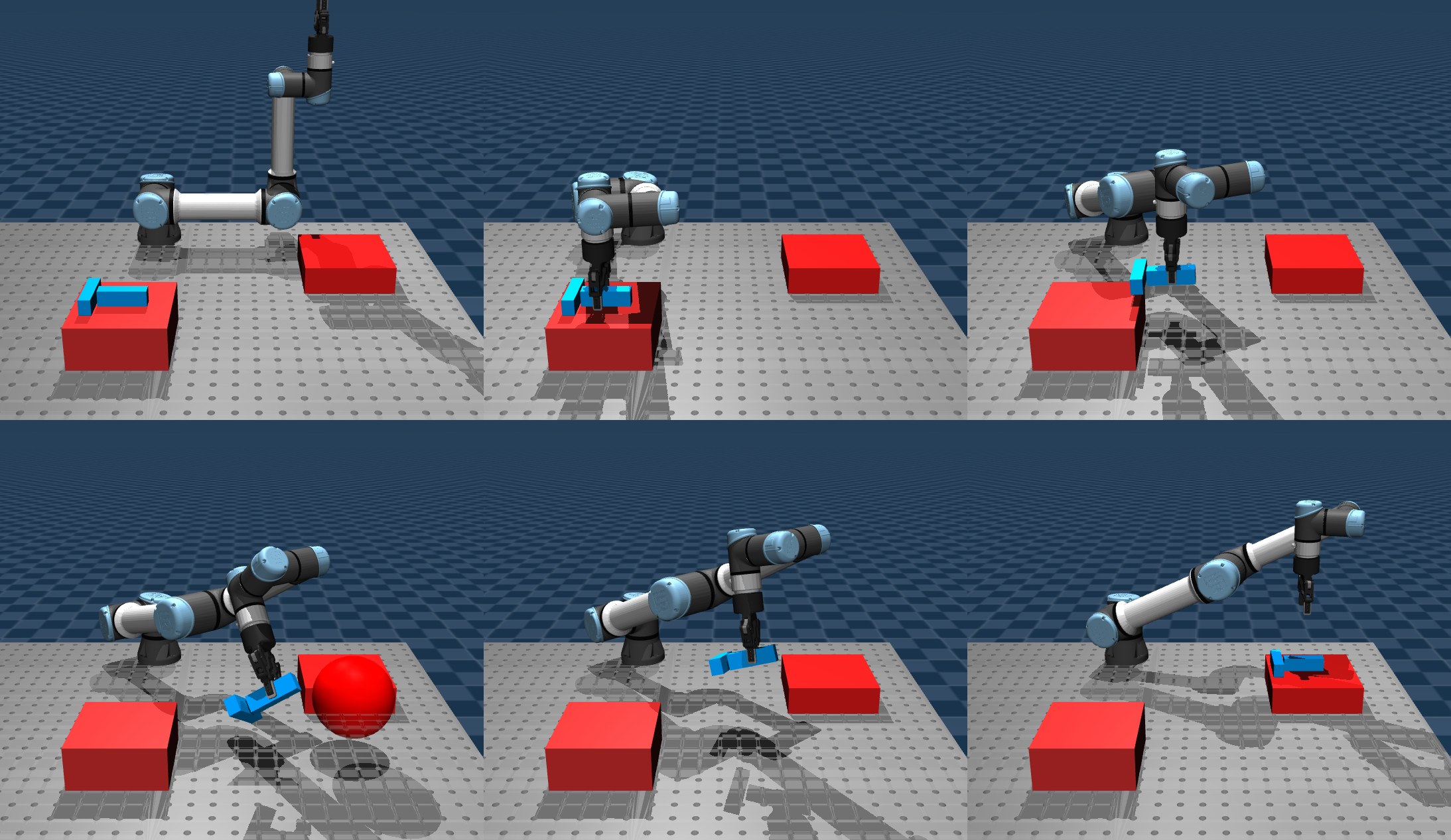}
\caption{In the MuJoCo simulation, PARRY redirects the end effector around a dynamic spherical obstacle while completing the transport task.}
\label{fig:sim}
\end{figure}

This section evaluates PARRY's ability to safely modify teleoperation commands at the end-effector velocity level. Because large-scale repetition of human-in-the-loop experiments under identical randomized scenarios is impractical, nominal end-effector velocities are generated from a quintic-polynomial joint-space pick-and-place trajectory as a substitute for online joystick commands. A PD controller continues to track the joint commands, while PARRY modifies only the nominal end-effector velocity without changing the underlying controller. This setup enables large-scale paired comparisons under identical task and obstacle conditions.

\subsection{Experimental Setup}

We construct a MuJoCo simulation environment comprising a UR5e manipulator, a Robotiq 2F-85 gripper, a workbench, and a manipulated object~\cite{todorov2012mujoco}. The manipulator performs a pick--transport--release task, as shown in Fig.~\ref{fig:sim}, while the safety module modifies only the end-effector velocity command and leaves the tracking controller unchanged. The physics simulation uses a time step of 2~ms, the velocity controller runs at 100~Hz, and the depth point cloud is updated at 30~Hz.

During the execution of the primary task, a spherical obstacle with a radius of 0.10~m is launched from a hemispherical surface approximately 1.5~m from the end effector. In each trial, the sphere speed is sampled uniformly from $[0.5,3.5]$~m/s, the launch time from $[0.5,6.0]$~s, and the target point from a spherical region of radius 0.25~m centered near the end effector. Zero-mean Gaussian noise with standard deviations of 0.01~m and 0.05~m/s is added to the point-cloud positions and obstacle velocity, respectively.

To emulate the obstacle-observation latency introduced by depth perception and communication pipelines in teleoperation, delays of $\tau\in\{0,100,150\}$~ms are applied to the obstacle point cloud, with $\tau=0$~ms serving as the reference condition. For each latency setting, 1,000 randomized scenarios are generated. Under the same random seed, all methods share identical sphere speeds, launch times, directions, and target points, enabling paired comparisons.

\subsection{Baselines and Evaluation Metrics}

The comparison includes three baselines and three ablations. APF-VS uses the same point-cloud repulsive-force formulation as PARRY but adds the repulsive force directly to the nominal task velocity. CBF-QP formulates point-cloud obstacle constraints as a kinematic CBF quadratic program with slack variables~\cite{ames2016control}. Related plug-and-play CBF-based safety layers have also been integrated with VLA policies to improve collision avoidance during robotic manipulation~\cite{hu2026vlsavisionlanguageactionmodelsplugandplay}. SSM scales the nominal task velocity according to the end-effector--obstacle distance and approach speed~\cite{marvel2017implementing}. Starting from the complete PARRY framework, we disable overshoot protection, TTC prediction, and relative velocity individually to obtain PARRY w/o Guard, PARRY w/o Pred, and PARRY w/o Rel, respectively. The complete PARRY framework incorporates relative velocity, TTC-based adaptive prediction, and overshoot protection.

\begin{table*}[t]
\centering
\caption{Main Monte Carlo Results over 1{,}000 Paired Trials per Method and Delay Condition}
\label{tab:main_results}

\renewcommand{\arraystretch}{1.15}
\setlength{\tabcolsep}{5.5pt}
\small

\begin{tabular}{
@{\hspace{5pt}}
wc{2.7cm}|
*{3}{wc{1.55cm}}|
*{3}{wc{1.55cm}}|
wc{1.8cm}
@{\hspace{5pt}}
}

\specialrule{1.05pt}{0pt}{1.0pt}
\specialrule{0.55pt}{0pt}{0pt}

\rule{0pt}{2.8ex}
\multirow{2}{*}{\textbf{Method}}
& \multicolumn{3}{c|}{\textbf{CAR (\%) $\uparrow$}}
& \multicolumn{3}{c|}{\textbf{Min clearance (m) $\uparrow$}}
& \multirow{2}{*}{\textbf{Time (ms) $\downarrow$}} \\

\cmidrule(lr){2-4}\cmidrule(lr){5-7}

& \textbf{0\,ms}
& \textbf{100\,ms}
& \textbf{150\,ms}
& \textbf{0\,ms}
& \textbf{100\,ms}
& \textbf{150\,ms}
& \\

\midrule[0.65pt]
\addlinespace[1pt]

APF-VS
& 79.1 & 74.7 & 73.7
& 0.176 & 0.142 & 0.136
& \textbf{0.024} \\

CBF-QP
& 76.2 & 72.3 & 71.9
& 0.151 & 0.125 & 0.124
& 0.729 \\

SSM
& 71.8 & 71.1 & 71.7
& 0.122 & 0.122 & 0.124
& 0.230 \\

\midrule[0.55pt]
\addlinespace[1pt]

PARRY w/o Rel
& 72.4 & 72.7 & 72.7
& 0.134 & 0.131 & 0.130
& 0.049 \\

PARRY w/o Pred
& 78.8 & 74.7 & 73.5
& 0.179 & 0.142 & 0.134
& 0.053 \\

PARRY w/o Guard
& 82.5 & 80.2 & \textbf{80.1}
& 0.215 & \textbf{0.184} & \textbf{0.165}
& 0.074 \\

\textbf{PARRY (ours)}
& \textbf{82.7} & \textbf{80.6} & \textbf{80.1}
& \textbf{0.222} & 0.183 & 0.164
& 0.101 \\

\specialrule{0.55pt}{1pt}{1.0pt}
\specialrule{1.05pt}{0pt}{0pt}
\end{tabular}

\vspace{3pt}
\parbox{\textwidth}{\footnotesize
\textbf{Notes:}
\textbf{CAR}: Collision Avoidance Rate;
\textbf{Time}: median per-cycle cost of the avoidance module.
}
\end{table*}

For a fair comparison, all methods share the same task trajectories, randomized obstacles, point-cloud observations, and underlying controller, and the key parameters of each method are calibrated independently. The primary metric is the end-effector collision avoidance rate (CAR). Secondary metrics include the median minimum clearance between the end effector and the sphere surface and the median collision-avoidance computation time. The effect of overshoot protection is quantified using large-angle flips in the repulsive-force direction and frames with an erroneous repulsive-force direction. A large-angle flip occurs when the angle of the repulsive force between two consecutive valid frames exceeds $90^\circ$, whereas an erroneous direction occurs when the obstacle is still approaching but the repulsive force points toward it.
CAR is computed over 1,000 trials in each group. Because all methods share paired scenarios, statistical significance is assessed using the exact McNemar test, with Holm correction applied to repeated comparisons across the three latency settings~\cite{mcnemar1947note, holm1979simple}.
\subsection{Results and Discussion}

\begin{figure}[t]
\centering
\includegraphics[width=\columnwidth]{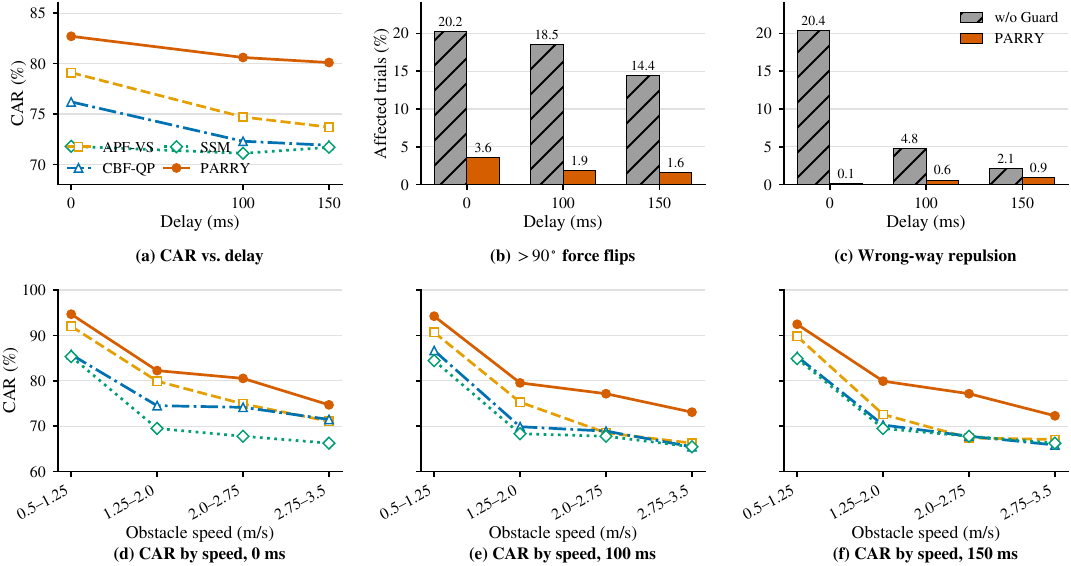}
\caption{Monte Carlo evaluation 
(a) CAR under increasing obstacle-observation delay.
(b) Percentage of trials containing at least one repulsive-force direction change greater than $90^\circ$ between consecutive valid frames.
(c) Percentage of trials containing at least one wrong-way frame, in which the obstacle is approaching while the repulsive force points toward it.
Panels (b) and (c) compare PARRY with and without the overshoot guard.
(d)--(f) CAR stratified by obstacle speed at 0, 100, and 150 ms, respectively.}
\label{fig:mc_panels}
\vspace{-6mm}
\end{figure}

The results are reported in Table~\ref{tab:main_results} and Fig.~\ref{fig:mc_panels}.

\subsubsection{Baseline Safety Comparison}

Under the baseline setting, PARRY reduces the end-effector collision rate by 17.2\%, 27.3\%, and 38.7\% relative to APF-VS, CBF-QP, and SSM, respectively. The paired McNemar tests remain significant after Holm correction, with $p<0.01$. These results show that PARRY achieves a lower end-effector collision rate than the three baselines.

\subsubsection{Robustness to Obstacle-Observation Latency}

As $\tau$ increases from 0 to 150~ms, the collision rates of APF-VS and CBF-QP increase by 5.4 and 4.3 percentage points (pp), respectively, whereas that of the complete PARRY framework increases by only 2.6~pp and remains the lowest under every latency setting. Across the three settings, PARRY reduces the collision rate by 17.2\%--24.3\%, 27.3\%--30.0\%, and 29.7\%--38.7\% relative to APF-VS, CBF-QP, and SSM, respectively. These paired improvements are statistically significant at every latency setting after Holm correction ($p<0.01$). PARRY therefore maintains its CAR advantage under obstacle-observation latency and degrades more gradually than APF-VS and CBF-QP.

\subsubsection{Ablation Study}

At 0~ms latency, PARRY w/o Pred achieves a 23.2\% relative reduction in the end-effector collision rate compared with PARRY w/o Rel, suggesting that relative-velocity processing in the obstacle reference frame better captures the dynamic relationship between the obstacle and end effector. After TTC-based point-cloud prediction is introduced, PARRY w/o Guard reduces the collision rate by 17.5\%, 21.7\%, and 24.9\% relative to PARRY w/o Pred at 0, 100, and 150~ms, respectively. Without prediction, the degradation with increasing latency is comparable to that of APF-VS. Prediction substantially reduces this degradation, with larger benefits at higher latencies. Although overshoot protection has only a limited effect on the collision rate, it suppresses reverse repulsive forces caused by predicted points passing beyond the end effector. Consequently, this substantially improves motion stability and operator experience during teleoperation. At 0~ms latency, the proportion of trials exhibiting large-angle flips decreases from 20.2\% to 3.6\%, while the proportion exhibiting erroneous repulsive-force directions decreases from 20.4\% to 0.1\%. 

\subsubsection{Safety Margin and Computational Cost}

PARRY achieves the largest median minimum end-effector clearance, at 0.188~m, compared with 0.150~m for APF-VS and 0.135~m for CBF-QP, indicating a larger safety margin than the baseline methods. Its median collision-avoidance computation time is 0.101~ms, substantially lower than the 0.729~ms required by CBF-QP and compatible with the 100~Hz control frequency used for teleoperation.
\begin{figure}[t]
\centering
\includegraphics[width=\columnwidth]{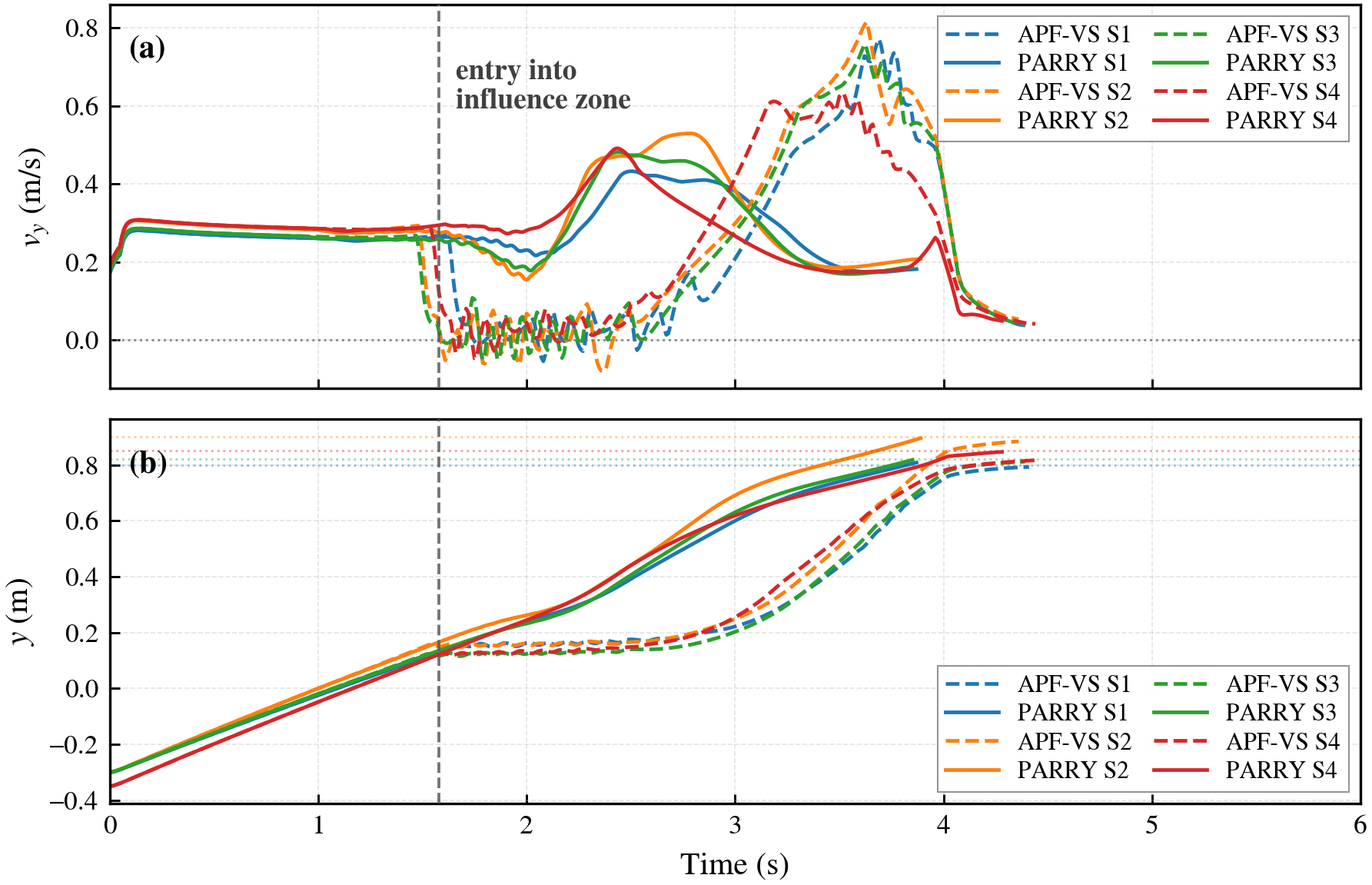}
\caption{End-effector forward velocity $v_y$ and position $y$ for APF-VS (dashed) and PARRY (solid) under four symmetric configurations.}
\label{fig:localmin}
\vspace{-6mm}
\end{figure}
\subsubsection{Local-Minimum Cases in Symmetric Two-Sphere Configurations}

To compare the velocity-superposition mechanism of APF-VS with the directional-redirection mechanism of PARRY, we further construct four static configurations in which APF methods are prone to local minima.

As shown in Fig.~\ref{fig:localmin}, the end-effector forward velocity $v_y$ and position $y$ exhibit consistent trends across all four configurations for APF-VS (dashed) and PARRY (solid). As the end effector approaches a local-minimum point, the APF-VS velocity $v_y$ persistently oscillates around zero, producing a pronounced plateau in $y$. By rotating only the direction of the relative velocity while preserving its magnitude, PARRY maintains the forward velocity component, traverses this region more smoothly, and reaches the target earlier.

These results indicate PARRY mitigates the transient stagnation caused by APF velocity superposition. It maintains a responsive end-effector motion under continuous teleoperation commands and reduces the risk that an operator misinterprets the stagnation as a hardware failure.

\section{Physical-Robot Experiments}

To evaluate the feasibility of PARRY in a physical environment and to complement the simulation experiments with real human-in-the-loop interaction, we construct a physical-robot validation platform. The complete experiments are provided in the supplementary video.

The hardware platform uses a 7-DoF Flexiv Rizon 4 manipulator, while the operator provides desired end-effector pose commands through a Meta Quest 3 controller. An Intel RealSense D435 depth camera mounted in front of the workbench captures the workspace point cloud at 30~Hz. The software pipeline uses TactAR~\cite{xue2025reactive} as the base teleoperation framework. When PARRY is enabled, point-cloud preprocessing, obstacle prediction, and velocity redirection are executed on a local workstation at a control frequency of 100~Hz. The corrected velocity command is then sent directly to the low-level controller of the manipulator.

\subsection{Task Setup and Performance Evaluation}

The operator is required to sequentially grasp four paper cups placed in a cluttered arrangement on one side of the workbench, stack them neatly on the other side, and then move the entire stack back to its original location. The cups vary in orientation and arrangement, and accurate alignment is required during stacking, motivating human-in-the-loop teleoperation rather than relying solely on fully autonomous VLA-based execution. Static and dynamic obstacles are introduced during task execution to evaluate the collision-avoidance response and motion behavior of PARRY.
\begin{figure}[htbp]
\centering
\includegraphics[width=\columnwidth]{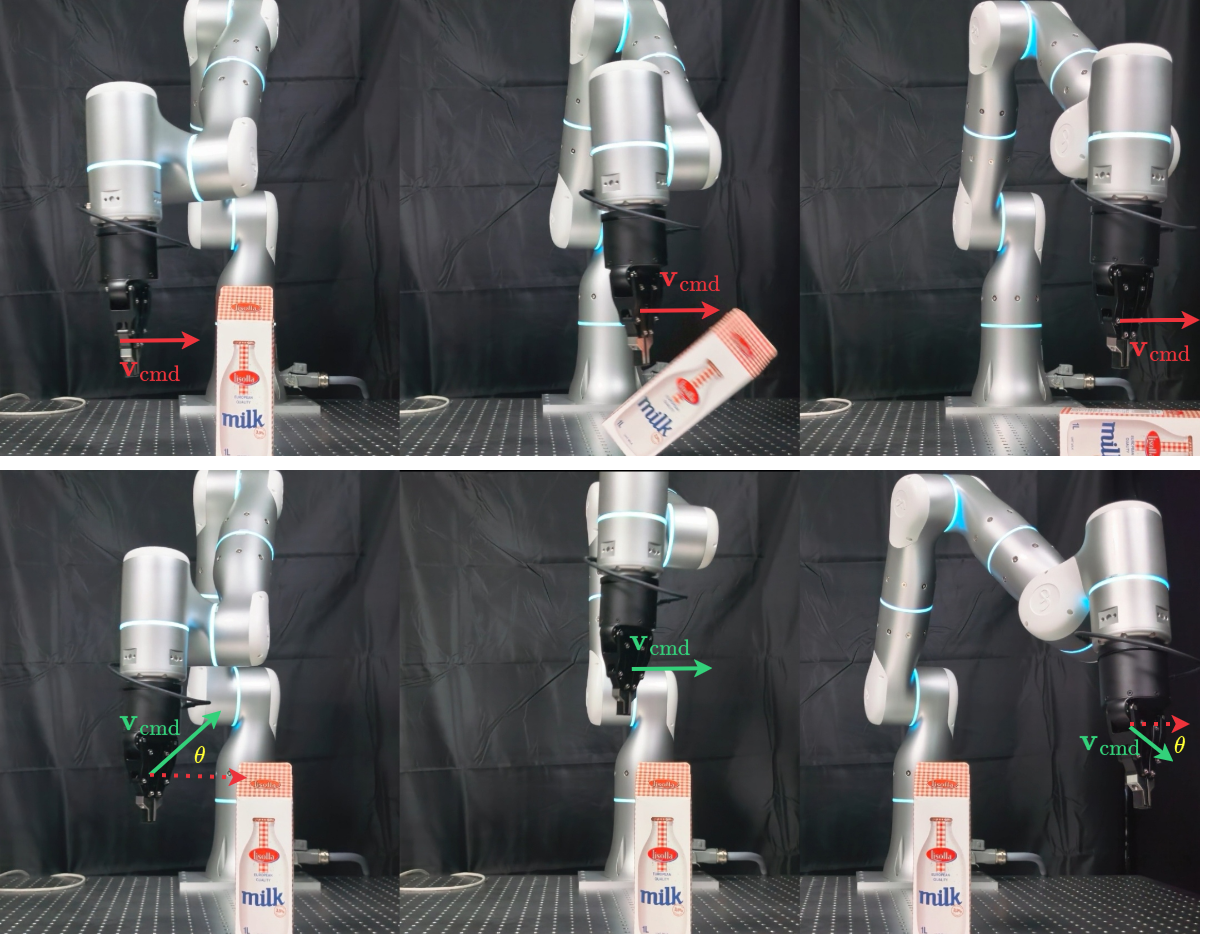}
\caption{Static obstacle scenario. \textbf{Top:} Baseline collision. \textbf{Bottom:} PARRY successfully steers around the obstacle in real time.}
\label{fig:teaser}
\vspace{-6mm}
\end{figure}

\begin{figure}[htbp]
\centering
\includegraphics[width=\columnwidth,trim=0 30 0 50,clip]{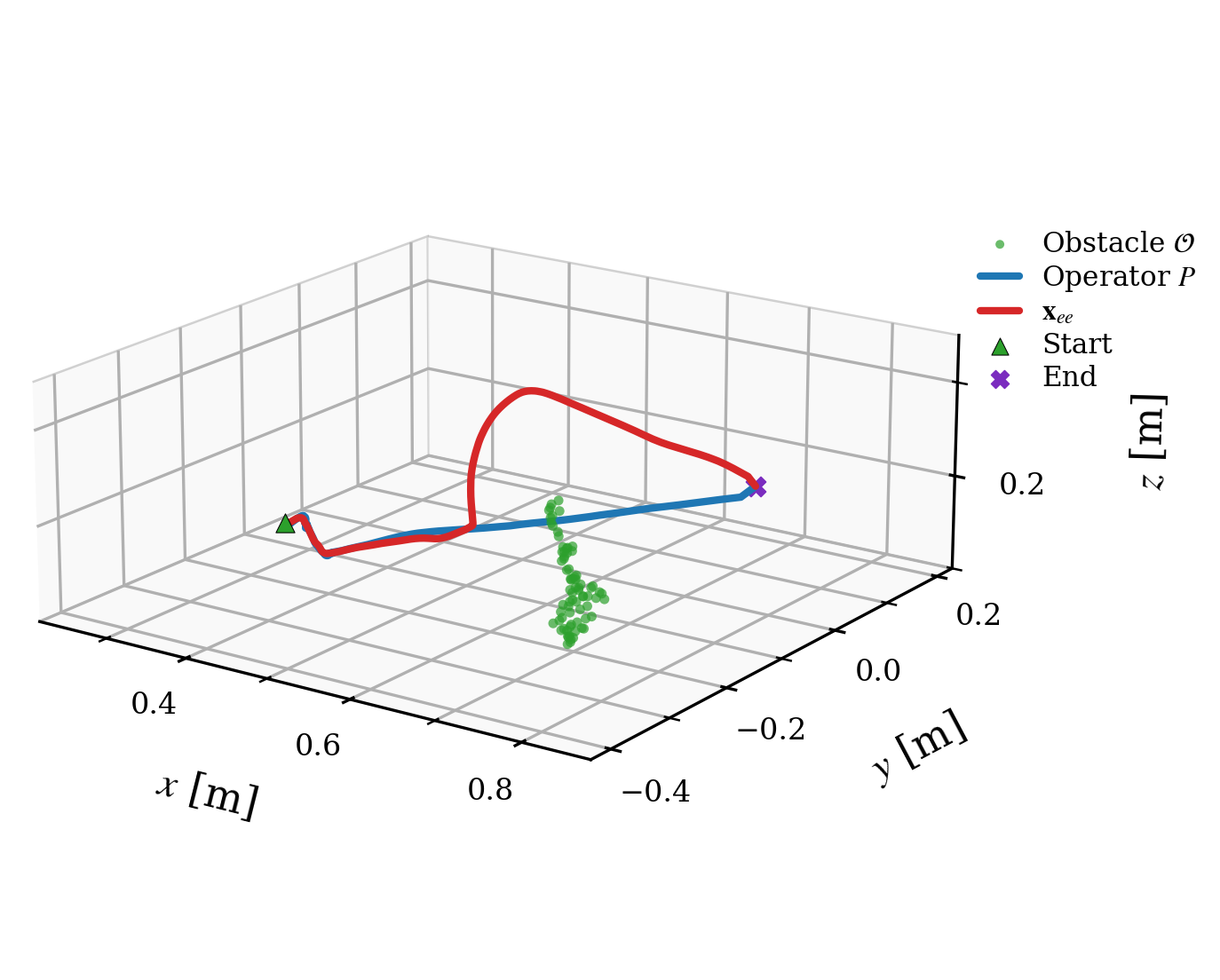}
\caption{Physical-robot trajectories.}
\label{fig:teleop}
\vspace{-6mm}
\end{figure}
\subsection{Evaluation in Static and Dynamic Scenarios}
 We evaluate PARRY under both static blockage and dynamic intrusions. In the static scenario (Fig.~\ref{fig:teaser}, Fig.~\ref{fig:teleop}), base teleoperation framework collides with an obstacle placed in the nominal path, whereas PARRY redirects the end-effector to bypass the blockage and successfully complete the cup-stacking task. In the dynamic scenario, PARRY successfully evades a moving obstacle manually introduced during transport in real time. These physical-robot experiments demonstrate PARRY's collision-avoidance effectiveness and implementation feasibility in the tested teleoperation setup.
\section{Conclusion}

This paper presented PARRY, a reactive predictive collision-avoidance framework for improving safety in teleoperation. PARRY used TTC-inspired look-ahead point-cloud reconstruction, redirected the relative velocity in the obstacle reference frame without changing its magnitude, and suppressed reverse repulsive forces through overshoot protection. Paired Monte Carlo simulations showed that PARRY achieved the highest or jointly highest safety rate under all three obstacle-observation latencies and mitigated transient APF stagnation in symmetric configurations. Physical-robot experiments further demonstrated its teleoperation collision-avoidance capability and deployment feasibility under static and dynamic obstacle scenarios. This study remained limited to end-effector collision avoidance. Future work will extend the collision-avoidance constraints to all manipulator links and evaluate the effects of PARRY on teleoperation efficiency and operator experience in human–robot interaction tasks.

\section*{Acknowledgments}
 We disclose using Gemini 3.1 Pro, ChatGPT and Opus 4.8 to assist with language polishing, creating conceptual figures, and supplementary video voiceovers. The authors take full responsibility for the final content.

\bibliographystyle{IEEEtran}
\let\oldbibliography\thebibliography
\renewcommand{\thebibliography}[1]{
  \oldbibliography{#1}
  \setlength{\itemsep}{-0.2pt} % 将条目间距设为0
}
\bibliography{refs}

\end{document}